\documentclass[11pt]{article}

\usepackage[final]{acl}

\usepackage{times}
\usepackage{latexsym}

\usepackage[T1]{fontenc}
\usepackage[utf8]{inputenc}

\usepackage{microtype}

\usepackage{inconsolata}

\usepackage{graphicx}

\usepackage{amsmath}
\usepackage{booktabs}
\usepackage{multirow}
\usepackage{dblfloatfix}
\usepackage{algorithm}
\usepackage{algpseudocode}
\usepackage{pifont}
\usepackage{amssymb}

\usepackage{tcolorbox}
\tcbuselibrary{breakable, skins}

\usepackage{tikz}

\definecolor{aclblue}{RGB}{207, 226, 243}
 \definecolor{aclcoral}{RGB}{210, 90, 48}

\title{IndexRAG: Index-Time Reasoning for Multi-Hop\\
Retrieval-Augmented Generation}

\author{
  \textbf{Zhenghua Bao},
  \textbf{Yi Shi}
\\
  Continuum AI
\\
  {
  \small
  \href{mailto:research@orcarouter.ai}{research@orcarouter.ai}
  }
}

\begin{document}
\maketitle
\begin{abstract}
Multi-hop question answering (QA) requires reasoning across multiple documents, yet existing retrieval-augmented generation (RAG) approaches address this either through graph-based methods requiring additional online processing or iterative multi-step reasoning. We present IndexRAG, a novel approach that shifts cross-document reasoning from online inference to offline indexing. IndexRAG identifies bridge entities shared across documents and generates bridging facts as independently retrievable units, requiring no additional training or fine-tuning. Experiments on three widely-used multi-hop QA benchmarks (HotpotQA, 2WikiMultiHopQA, MuSiQue) show that IndexRAG improves F1 over Naive RAG by 4.6 points on average, while requiring only single-pass retrieval and a single LLM call at inference time. When combined with IRCoT, IndexRAG achieves the best average performance among all evaluated methods, including graph-based baselines such as HippoRAG2 and FastGraphRAG, while relying on a flat vector index. Our code is available at \url{https://github.com/Continuum-AI-Corp/IndexRAG}.
\end{abstract}

\begin{table*}[h]
\centering
\begin{tabular}{lccccc}
\toprule
& \textbf{Single-pass} & \textbf{Cross-doc} & \textbf{Single LLM} & \textbf{Training-} & \textbf{Index-time} \\
& \textbf{retrieval} & \textbf{reasoning} & \textbf{call} & \textbf{free} & \textbf{reasoning} \\
\midrule
Naive RAG        & \checkmark & \texttimes & \checkmark & \checkmark & \texttimes \\
HippoRAG2         & \texttimes & \checkmark & \texttimes & \checkmark & \texttimes \\
IRCoT            & \texttimes & \checkmark & \texttimes & \checkmark & \texttimes \\
\midrule
\textbf{IndexRAG (Ours)} & \checkmark & \checkmark & \checkmark & \checkmark & \checkmark \\
\bottomrule
\end{tabular}
\caption{Qualitative comparison of different RAG approaches. IndexRAG is the only method that achieves cross-document reasoning with single-pass retrieval and a single LLM call at inference time.}
\label{tab:comparison}
\end{table*}

\section{Introduction}

Large language models (LLMs), built on the Transformer architecture \cite{vaswani2017attention}, have achieved remarkable performance across a wide range of tasks, spanning language \cite{JMLR:v21:20-074, brown2020language}, vision \cite{dosovitskiy2021image, pmlr-v139-radford21a}, and speech \cite{zhang-etal-2023-speechgpt, defossez2024moshi}. However, their reliance on static parametric knowledge limits their ability to incorporate domain-specific or up-to-date information, often leading to hallucinations \cite{huang2025survey}. Retrieval-Augmented Generation (RAG) mitigates this by grounding generation in an external knowledge base, allowing the model to retrieve relevant context before responding \cite{lewis2020retrieval}.

\begin{figure}[t]
  \includegraphics[width=\linewidth, trim=135 10 90 10, clip]{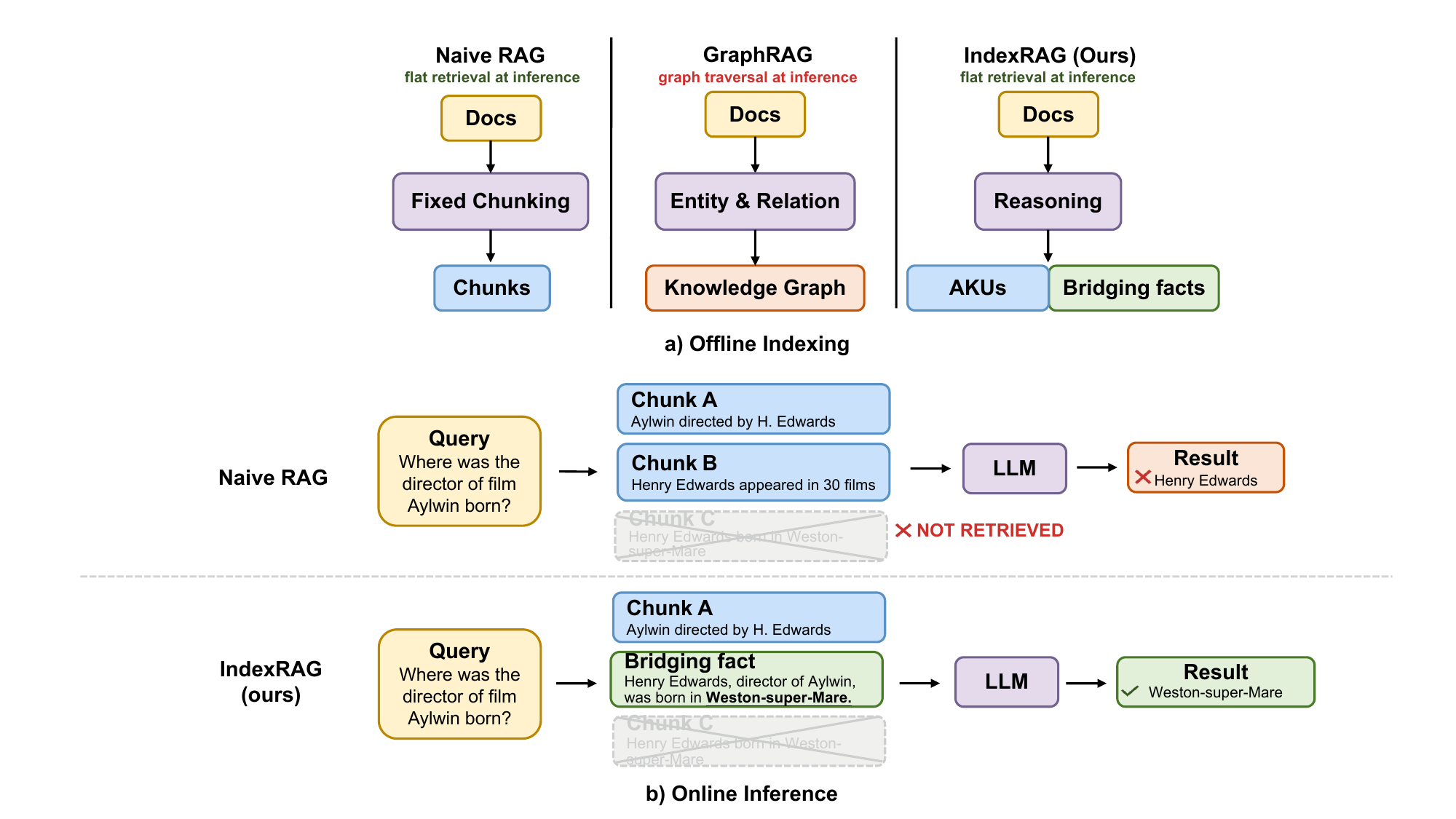}
  \caption{(a) Comparison of different RAG approaches during offline indexing. IndexRAG shifts reasoning to index time, generating bridging facts stored alongside AKUs in a unified vector store. (b) Example from 2WikiMultiHopQA~\cite{ho2020constructing}. The query \textbf{``Where was the director of the film Aylwin born?''} requires two-hop reasoning. With naive RAG, \colorbox{gray!20}{Chunk C} containing the director's birthplace is not retrieved, causing the LLM to output the director's name instead. With IndexRAG, a \colorbox{green!15}{bridging fact} supplying the director's birthplace is retrieved directly, completing the reasoning chain. The panel is illustrative, with the retrieved units reported in Table~\ref{tab:case-study} and analyzed in Section~\ref{sec:case-study}.}
  \label{fig:motivation}
\end{figure}

While effective for single-hop questions, conventional RAG pipelines retrieve passages independently and struggle when answering a question requires synthesizing information across multiple documents. Multi-hop QA requires reasoning over multiple pieces of evidence spread across different documents to reach the correct answer~\cite{yang-etal-2018-hotpotqa, ho2020constructing, trivedi-etal-2022-musique}. Graph-based RAG systems~\cite{edge2024local, gutierrez2024hipporag,  pmlr-v267-gutierrez25a, you-etal-2025-ms} address this by constructing knowledge graphs that explicitly represent inter-document relationships. However, these methods often require multi-step online processing, such as query-time entity extraction, graph traversal, and multiple LLM calls, which increases inference cost and latency. Iterative approaches such as IRCoT~\cite{trivedi2023interleaving} decompose complex queries through multiple rounds of retrieval and generation, resulting in higher inference cost and slower response times.

To address these limitations, we propose IndexRAG, a novel approach that
shifts cross-document reasoning from online inference to offline indexing.
We observe that these connections do not depend on the query. When two
documents mention the same entity, the link between them is already fixed by
what the documents say. This makes it possible
to precompute these reasoning connections at indexing time, transforming 
per-query reasoning into an offline step. We refer to this paradigm as 
\textit{index-time reasoning}. During indexing, we extract atomic 
knowledge units (AKUs) and entities from each document, and identify 
entities that appear across multiple documents. We then prompt an LLM 
to generate \textit{bridging facts} that capture cross-document reasoning
by linking related evidence from different sources. Both AKUs and bridging facts are stored in a unified vector store.
Figure~\ref{fig:motivation} sets this against existing pipelines and traces one
query through both.
At inference time, a single retrieval pass and a single LLM call suffice, with no
graph traversal or repeated retrieval rounds. Table~\ref{tab:comparison}
summarizes how IndexRAG differs from existing approaches.

Our main contributions are as follows:
\begin{itemize}
    \item We propose \textbf{IndexRAG}, a novel approach that shifts 
    cross-document reasoning from online inference to offline indexing.
    \item We introduce \textit{bridging facts}, a new type of retrieval 
    unit that encodes cross-document reasoning as independently 
    retrievable entries in a flat vector store, along with a balanced 
    context selection mechanism to control their proportion at retrieval time.
    \item We propose a training-free framework that is agnostic to the underlying retrieval strategy and compatible with iterative methods such as IRCoT, requiring no fine-tuning of the embedding model or LLM.
    \item Experiments on three multi-hop QA benchmarks show IndexRAG 
outperforms all single-call baselines on average and, when combined 
with IRCoT, achieves the best average performance among all evaluated methods.
\end{itemize}

\section{Related Work}

\subsection{Retrieval-Augmented Generation}
To mitigate hallucinations arising from the static parametric knowledge of LLMs, RAG augments generation with external retrieval \cite{lewis2020retrieval}. In a typical RAG pipeline, documents are encoded into a vector store, and relevant passages are retrieved based on embedding similarity at inference time. 
Recent research has focused on improving retrieval representations. \citet{karpukhin2020dense} replace sparse retrieval with learned dense representations, while \citet{gao2023precise} prompt the LLM to generate a hypothetical answer and use it as the retrieval query to better align with target passages.
Other work investigates how documents should be segmented for indexing. \citet{chen-etal-2024-dense} show that finer-grained units such as propositions yield stronger retrieval than fixed-size chunks. While propositions decompose individual passages into finer units, they do not capture reasoning across documents. RAPTOR \cite{sarthi2024raptor} recursively summarizes passages into a hierarchical tree structure for retrieval.

These methods primarily focus on improving how individual passages are represented or organized. In contrast, our work focuses on the implicit reasoning connections across documents, generating new retrieval units that make cross-document inferences explicitly retrievable.

\begin{figure*}[!htbp]
  \centering
  \includegraphics[width=\linewidth, trim=30 150 20 90, clip]{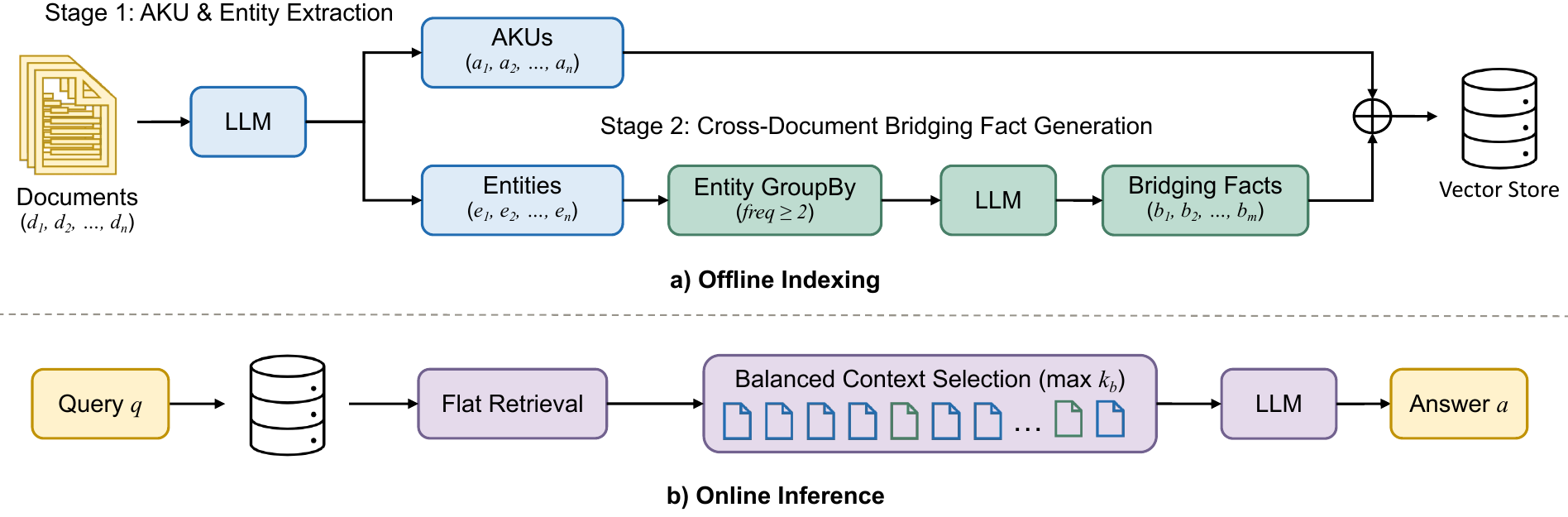}
  \caption{Overview of IndexRAG. (a) Offline Indexing: In Stage 1, the LLM extracts AKUs and associated entities from each document. In Stage 2, documents sharing the same entity are identified and used to generate bridging facts. The resulting AKUs and bridging facts are encoded and stored in a unified vector store. (b) Online Inference: the user query $q$ is encoded and used to retrieve relevant context. After applying balanced context selection with $k_b$, the selected context (a mix of AKUs and bridging facts) is fed to the LLM to generate the answer $a$.}
  \label{fig:overview}
\end{figure*}

\subsection{Multi-hop Reasoning in RAG}
Standard RAG retrieves passages independently, struggling with 
questions that require reasoning across multiple documents 
\cite{yang-etal-2018-hotpotqa, ho2020constructing, trivedi-etal-2022-musique}. 
Two main directions have emerged to address this limitation.

The first incorporates graph structures into RAG. GraphRAG \cite{edge2024local} constructs an entity graph from the corpus, applies community detection, and generates summaries for each community to support global reasoning. FastGraphRAG\footnote{\url{https://github.com/circlemind-ai/fast-graphrag}}~is another graph-based approach that prioritizes inference efficiency. HippoRAG \cite{gutierrez2024hipporag} draws inspiration from the human hippocampal memory system, using a knowledge graph as a long-term memory index to connect related passages, with Personalized PageRank for graph-based retrieval. HippoRAG2 \cite{pmlr-v267-gutierrez25a} extends this line with additional fact-level embeddings while retaining Personalized PageRank traversal at query time. MS-RAG \cite{you-etal-2025-ms} maintains separate vector stores for entities, relations, and chunks, using vector search for entity extraction, but still requires graph traversal at inference time. Although these methods improve multi-hop performance, they require 
additional inference-time processing, such as entity extraction, 
graph traversal, or reranking, beyond standard vector search. Recent agentic memory systems~\citep{chhikara2025mem0,kang2025memory,NEURIPS2025_19909c36,hu-etal-2026-evermemos} focus on memory construction for agentic or dialogue settings, while our work studies index-time reasoning for fixed-corpus multi-hop QA.

The second direction uses iterative retrieval. IRCoT \cite{trivedi2023interleaving} interleaves chain-of-thought reasoning with retrieval, using intermediate reasoning steps to formulate new queries across multiple rounds. This improves recall of relevant evidence but introduces repeated retrieval and generation cycles that increase inference latency. Self-Correcting RAG \cite{xu-etal-2026-self} pairs a knapsack formulation of context selection with an NLI-guided tree search that can issue further retrieval calls while exploring reasoning trajectories.

In contrast, IndexRAG shifts cross-document reasoning to indexing 
time, making implicit connections retrievable through standard 
vector search without additional inference-time overhead.

\section{Method}

\subsection{Overview}

IndexRAG follows a two-phase pipeline: offline indexing and online inference.
During offline indexing, we first extract atomic knowledge units (AKUs) 
and entities from each document (\textbf{Stage 1}), then generate 
bridging facts that capture cross-document reasoning by linking 
documents through shared entities (\textbf{Stage 2}).
Both AKUs and bridging facts are encoded and stored in a unified 
vector store, making cross-document reasoning directly accessible 
through standard vector search.
During online inference, we perform single-pass retrieval with balanced 
context selection and prompt the LLM for answer generation.
The overall pipeline is illustrated in Figure~\ref{fig:overview} and detailed below.

\subsection{Offline Indexing}

\subsubsection{Stage 1: AKU and Entity Extraction}

Given a corpus of $n$ documents $\mathcal{D} = \{d_1, d_2, \dots, d_n\}$, we prompt an LLM to extract a set of atomic facts, structured as question-answer pairs, and associated entities from each document $d_i$.
We retain only the answers, merging them into a single text unit 
per document as the minimal retrievable unit. We refer to this unit as an \textit{atomic knowledge unit} (AKU), denoted $a_i$.
The LLM simultaneously extracts a set of entities $E_i = \{e_1^{(i)}, e_2^{(i)}, \dots\}$. Each AKU $a_i$~is then encoded by a dense embedding model $f$ and stored in a flat vector store $\mathcal{V}$.
Empirically, the choice of extraction method in Stage 1 has a limited effect on overall pipeline performance (Section~\ref{sec:stage_1}). Nevertheless, the AKU representation aggregates atomic facts rather than preserving raw passages with unrelated context, providing finer-grained retrieval units similar to propositions~\cite{chen-etal-2024-dense}.

\subsubsection{Stage 2: Bridging Fact Generation}

The key observation motivating this stage is that documents sharing 
common entities often contain complementary information. While each 
document alone may only provide partial evidence, their combination 
can yield reasoning conclusions not explicitly stated in any single 
source. By making these implicit connections explicit and retrievable, 
Stage 2 enables cross-document reasoning without any additional 
inference-time processing.

\paragraph{Bridge Entity Identification.}
We aggregate all entities from Stage 1 and identify \textit{bridge entities} that appear across multiple documents:
\begin{equation}
    \mathcal{E}_{\text{bridge}} = \left\{ e \in \bigcup_{i=1}^{n} E_i \ \middle|\ 2 \leq \text{df}(e) \leq \tau \right\}
\end{equation}
where $\text{df}(e) = |\{d_i \in \mathcal{D} : e \in E_i\}|$ is the document frequency of entity $e$, and $\tau$ is an upper-bound threshold.
The lower bound ensures that bridge entities connect at least two documents, while the upper bound $\tau$ excludes overly generic entities (e.g., common nouns or high-frequency terms) that would produce trivial bridging facts. 

% In practice, we set $\tau = 10$ and the sensitivity analysis is provided in Section~\ref{sec:implementation}.

\paragraph{Bridging Fact Generation.}
For each bridge entity $e \in \mathcal{E}_{\text{bridge}}$, let $\mathcal{D}_e = \{d_i \in \mathcal{D} : e \in E_i\}$ be the set of documents mentioning $e$.
We collect the subset of facts in each AKU $a_i$ that mention entity $e$, denoted $a_i[e]$, identified via string matching against the entity list from Stage~1.
We then prompt the LLM to generate bridging facts:
\begin{equation}
    \mathcal{B}_e = \text{LLM}\left(e,\ \{a_i[e] : d_i \in \mathcal{D}_e\}\right)
\end{equation}
where $\mathcal{B}_e$ is the resulting set of bridging facts.
We limit the number of source documents to 5 and facts per document to 8 to manage input length.

Unlike cross-document summaries, which compress existing content into shorter passages, bridging facts are constructed to directly answer implicit cross-document questions. Figure~\ref{fig:motivation}(b) illustrates this. Given Doc A stating ``Aylwin is directed by Henry Edwards''
and Doc B stating ``Henry Edwards was born in 
Weston-super-Mare'', IndexRAG generates the bridging fact 
``The director of the film Aylwin was born in 
Weston-super-Mare''. Unlike either source document, this 
fact directly encodes the two-hop reasoning chain and is 
semantically aligned with queries like ``Where was the 
director of Aylwin born?'', making it retrievable via 
standard vector search even though neither source document alone is sufficient 
to answer the query.

All bridging facts $\mathcal{B} = \bigcup_{e \in \mathcal{E}_{\text{bridge}}} \mathcal{B}_e$ are individually encoded by $f$ and stored alongside the AKUs in $\mathcal{V}$.
When a new document $d_{\text{new}}$ is added, only Stage~1 for $d_{\text{new}}$ and Stage~2 for affected bridge entities need to be re-executed. This includes both existing bridge entities that appear in $d_{\text{new}}$ ($E_{\text{new}} \cap \mathcal{E}_{\text{bridge}}$) and newly formed bridge entities whose document frequency reaches 2 after the addition. The rest of the index remains unchanged.

\subsection{Online Inference}
Given a query $q$, we encode it with the same embedding model $f$ and retrieve the top-$k$ entries from $\mathcal{V}$ by cosine similarity, where $v_j \in \mathcal{V}$ is either an AKU or a bridging fact.

\paragraph{Balanced Context Selection.}
The retrieved top-$k$ set typically contains a mix of AKUs and bridging facts.
However, because bridging facts are considerably shorter than AKUs on average (166 vs.\ 634 characters in our experiments), they tend to dominate the top-$k$, crowding out the longer AKUs.

To control this, let $R = \{r_1, r_2, \dots\}$ be all entries in $\mathcal{V}$ sorted by descending similarity to $q$, where each $r_j$ is either an AKU ($r_j \in \mathcal{A}$) or a bridging fact ($r_j \in \mathcal{B}$).
We greedily build the context $C$ by iterating over $R$ in order. Each entry is included if it is an AKU or if the number of bridging facts already in $C$ is below $k_b$, until $|C| = k$.
The procedure is summarized in Appendix~\ref{app:algorithm}.

The selected context $C$ is concatenated in similarity order and provided to the LLM for answer generation (see Figure~\ref{fig:overview}b), without any additional inference-time overhead.

\section{Experimental Setup}

\subsection{Datasets}

We evaluate on three widely-used multi-hop QA datasets: \textbf{HotpotQA} \cite{yang-etal-2018-hotpotqa}, \textbf{2WikiMultiHopQA} \cite{ho2020constructing}, and \textbf{MuSiQue} \cite{trivedi-etal-2022-musique}.

HotpotQA contains 113k questions requiring reasoning over multiple Wikipedia passages.
2WikiMultiHopQA provides multi-hop questions across different types (comparison, bridge comparison, compositional, and inference) with structured reasoning paths.
MuSiQue consists of 25k 2-4 hop questions composed from single-hop sub-questions, and is generally considered the most challenging among the three.
Following prior work \cite{press2023measuring, trivedi2023interleaving, gutierrez2024hipporag, pmlr-v267-gutierrez25a}, we sample 1,000 questions from each validation set and collect all associated passages (supporting and distractor) as the retrieval corpus. Dataset statistics are summarized in Table~\ref{tab:datasets}.

\begin{table}[t]
  \centering
  \begin{tabular}{lcc}
    \hline
    \textbf{Dataset} & \textbf{Questions} & \textbf{Passages} \\
    \hline
    HotpotQA & 1,000 & 9,827 \\
    2WikiMultiHopQA & 1,000 & 6,262 \\
    MuSiQue & 1,000 & 9,723 \\
    \hline
  \end{tabular}
  \caption{Dataset statistics for the 1,000 question subsets used in our experiments.}
  \label{tab:datasets}
\end{table}

\subsection{Implementation Details}

We use GPT-4o-mini~\footnote{\url{https://openai.com/index/gpt-4o-mini-advancing-cost-efficient-intelligence/}} as the LLM for all stages and all methods, including baselines, to ensure a fair comparison.
All text is encoded using OpenAI's text-embedding-3-small~\footnote{\url{https://developers.openai.com/api/docs/models/text-embedding-3-small}}, and we use FAISS \cite{douze2025faiss} for vector indexing. Prompt templates for all stages and answer generation are provided 
in Appendix~\ref{sec:prompts}.

\paragraph{Offline Indexing.}
In Stage 1, we pass each full document to the LLM for AKU and entity extraction.
In Stage 2, we set the entity frequency threshold $\tau = 10$, with a maximum of 5 source documents and 8 facts per document for each bridge entity. The offline pipeline requires one LLM call per document for Stage~1 and one LLM call per bridge entity for Stage~2, with the total Stage~2 cost proportional to $|\mathcal{E}_{\text{bridge}}|$. Entities with no meaningful cross-document connection are discarded, accounting for the non-empty rates in Table~\ref{tab:bridging-stats}. For baseline methods that require chunking (Naive RAG, RAPTOR, etc.), we split documents into passages of approximately 100 words following \citet{chen-etal-2024-dense}, with an 80-character overlap in our implementation.

\paragraph{Online Inference.}
All methods retrieve 20 candidates and select the top $k = 10$ for answer generation.
For IndexRAG, balanced context selection with $k_b = 3$ is additionally applied to control 
the proportion of bridging facts in the selected context. Unless otherwise specified,
this value is used throughout rather than tuned per dataset.

Since all evaluated datasets expect short-form answers (typically a few words), we set 
the generation temperature to 0 and max tokens to 50. For IRCoT experiments, we adopt the
interleaving procedure of \citet{trivedi2023interleaving}, with 3 reasoning steps
and top-20 retrieval per step chosen for our setup.
When combined with IndexRAG, bridging facts are included in the 
retrieval corpus and balanced context selection is applied at the 
final answer generation step, with $k_b = 3$.

\begin{table}[t]
  \centering
  \small
  \begin{tabular}{lccc}
    \hline
    & \textbf{HotpotQA} & \textbf{2Wiki} & \textbf{MuSiQue} \\
    \hline
    Bridge entities & 8,878 & 4,817 & 6,471 \\
    Bridging facts & 12,608 & 6,821 & 8,010 \\
    Non-empty rate & 85\% & 88\% & 79\% \\
    \hline
  \end{tabular}
  \caption{Bridging fact generation statistics.}
  \label{tab:bridging-stats}
\end{table}

\begin{table*}[t]
  \centering
  \small
  \begin{tabular}{l @{\hspace{6pt}} ccc @{\hspace{10pt}} ccc @{\hspace{10pt}} ccc @{\hspace{6pt}} | @{\hspace{6pt}} ccc}
    \hline
    & \multicolumn{3}{c}{\textbf{HotpotQA}} & \multicolumn{3}{c}{\textbf{2WikiMultiHopQA}} & \multicolumn{3}{c}{\textbf{MuSiQue}} & \multicolumn{3}{c}{\textbf{Average}} \\
    & EM & Acc & F1 & EM & Acc & F1 & EM & Acc & F1 & EM & Acc & F1 \\
    \hline
    BM25 & 47.2 & 51.2 & 60.3 & 31.6 & 32.5 & 35.9 & 9.8 & 10.7 & 19.2 & 29.5 & 31.5 & 38.5 \\
    Naive RAG & 50.2 & 54.0 & 63.6 & 42.2 & 44.5 & 47.7 & 19.0 & 20.4 & 29.9 & 37.1 & 39.6 & 47.1 \\
    FastGraphRAG & 50.0 & 53.8 & 63.5 & \textbf{49.5} & \textbf{54.8} & \textbf{57.4} & 17.5 & 18.8 & 27.2 & 39.0 & 42.5 & 49.4 \\
    RAPTOR & 50.2 & 54.1 & 63.6 & 42.3 & 44.2 & 47.8 & 19.3 & 20.3 & 29.7 & 37.3 & 39.5 & 47.0 \\
    \hline
    \textbf{IndexRAG} & \textbf{53.9} & \textbf{59.3} & \textbf{68.9} & 44.8 & 50.3 & 51.7 & \textbf{22.4} & \textbf{24.4} & \textbf{34.4} & \textbf{40.4} & \textbf{44.7} & \textbf{51.7} \\
    \hline
    \color{gray} HippoRAG2 & \textbf{\color{gray}56.5} & \textbf{\color{gray}60.7} & \textbf{\color{gray}70.5} & \color{gray}50.2 & \color{gray}55.7 & \color{gray}57.2 & \color{gray}23.8 & \color{gray}25.5 & \color{gray}34.7 & \color{gray}43.5 & \color{gray}47.3 & \color{gray}54.1 \\
    \color{gray} IRCoT & \color{gray}49.8 & \color{gray}54.9 & \color{gray}62.5 & \color{gray}39.7 & \color{gray}41.3 & \color{gray}43.1 & \color{gray}11.1 & \color{gray}12.3 & \color{gray}19.3 & \color{gray}33.5 & \color{gray}36.2 & \color{gray}41.6 \\
    \color{gray} IRCoT + IndexRAG & \color{gray}54.5 & \color{gray}59.7 & \color{gray}68.7 & \textbf{\color{gray}53.4} & \textbf{\color{gray}61.2} & \textbf{\color{gray}61.2} & \textbf{\color{gray}24.6} & \textbf{\color{gray}26.5} & \textbf{\color{gray}35.0} & \textbf{\color{gray}44.2} & \textbf{\color{gray}49.1} & \textbf{\color{gray}55.0} \\
    \hline
  \end{tabular}
  \caption{Multi-hop QA performance (\%) on three multi-hop QA benchmarks. Best single-call results per dataset are in \textbf{bold}. Best multi-call results are in \textbf{\color{gray}bold gray}. Methods requiring multiple LLM calls at inference time are greyed out.}
  \label{tab:main-results}
\end{table*}

\subsection{Baselines}

We compare IndexRAG against baselines spanning five categories:
(1)~\textbf{Sparse retrieval:} BM25 \cite{robertson1994some},
(2)~\textbf{Dense retrieval:} Naive RAG with fixed-size chunking using the same embedding model,
(3)~\textbf{Iterative approach:} IRCoT \cite{trivedi2023interleaving},
(4)~\textbf{Graph-based:} FastGraphRAG~\footnote{\url{https://github.com/circlemind-ai/fast-graphrag}} and HippoRAG2 \cite{pmlr-v267-gutierrez25a}, and
(5)~\textbf{Hierarchical:} RAPTOR \mbox{\cite{sarthi2024raptor}}.
We adopt FastGraphRAG as a representative graph-based baseline due to its improved efficiency over GraphRAG \cite{edge2024local} while maintaining competitive performance. HippoRAG2 embeds the query to retrieve candidate triples, filters them with an LLM, and then runs Personalized PageRank over the graph, resulting in additional online processing beyond flat vector search. Note that IRCoT was originally evaluated with BM25, so we re-run it with the dense retriever used throughout our experiments to keep the comparison controlled.
To demonstrate the generality of Stage~2, we also report IRCoT combined with bridging facts.

\subsection{Evaluation Metrics}

We evaluate using three metrics: Exact Match (EM), Accuracy (Acc), 
and F1 score, all computed after answer normalization following 
\citet{rajpurkar-etal-2016-squad}. In addition, we report average retrieval 
latency and the number of LLM calls at inference time to assess 
efficiency. Formal definitions are provided in 
Appendix~\ref{app:eval_metrics}.

\section{Results}

%19.2 29.8 46.6 59.5 41.8 55.0

% ============================================================
% TABLE 2: IRCoT + Bridging (Stage 2 Generality)
% ============================================================

% \begin{table}[t]
%   \centering
%   \small
%   \begin{tabular}{lccc}
%     \hline
%     \textbf{Method} & \textbf{HotpotQA} & \textbf{2Wiki} & \textbf{MuSiQue} \\
%     \hline
%     IRCoT & 49.8 & 47.8 & 22.1 \\
%     IRCoT + IndexRAG & \textbf{54.5} & \textbf{53.4} & \textbf{24.6} \\
%     $\Delta$ & +4.7 & +5.6 & +2.5 \\
%     \hline
%   \end{tabular}
%   \caption{Effect of adding bridging facts to IRCoT (EM, \%). Stage 2 consistently improves IRCoT across all three datasets.}
%   \label{tab:ircot-bridging}
% \end{table}

\subsection{Quantitative Results}
\subsubsection{Multi-hop QA Performance}
Table~\ref{tab:main-results} presents the main results across the three multi-hop QA benchmarks. We report F1 score as the primary metric as it captures partial credit for partially correct answers.

\paragraph{Single-call methods.}
Among methods requiring only a single LLM call at inference time, IndexRAG achieves the highest average scores across all three metrics. IndexRAG obtains an average F1 of 51.7, improving over Naive RAG (+4.6), BM25 (+13.2), FastGraphRAG (+2.3), and RAPTOR (+4.7). On HotpotQA and MuSiQue, IndexRAG ranks first across all metrics, with particularly strong gains on MuSiQue (34.4 vs.\ 29.9 for Naive RAG), where all baselines achieve their lowest scores across the three datasets. On 2WikiMultiHopQA, FastGraphRAG achieves the best single-call performance (57.4). Comparison and bridge comparison questions, which account for 44\% of this dataset, require parallel entity lookups that align naturally with explicit graph structures. IndexRAG's bridging facts are generated along sequential reasoning paths and provide limited benefit for these question types, as further analyzed in Section~\ref{sec:question_type}. Despite this, 
IndexRAG achieves the best average F1 among all single-call methods.

\begin{table}[t]
  \centering
  \small
  \begin{tabular}{lccc}
    \hline
    \textbf{Method} & \textbf{EM} & \textbf{Time(s)} & \textbf{Calls} \\
    \hline
    Naive RAG & 19.0 & 0.29 & 1.0 \\
    FastGraphRAG & 17.5 & 2.55 & 1.0 \\
    RAPTOR & 19.3 & 0.47 & 1.0 \\
    \hline
    \textbf{IndexRAG} & \textbf{22.4} & 0.30 & 1.0 \\
    \hline
    \color{gray} HippoRAG2 & \color{gray}23.8 & \color{gray}3.13 & \color{gray}2.0 \\
    \color{gray} IRCoT + IndexRAG & \color{gray}24.6 & \color{gray}1.08 & \color{gray}3.2 \\
    \hline
  \end{tabular}
  \caption{Quality-efficiency comparison (\%) on MuSiQue. \textcolor{gray}{Gray rows} indicate multi-call methods.}
  \label{tab:efficiency}
\end{table}

\begin{table*}[t]
  \centering
  \small
  \begin{tabular}{p{2.2cm} p{12.5cm}}
    \hline
    \multicolumn{2}{l}{\textbf{Query:} Where was the director of the film Aylwin born?} \\
    \multicolumn{2}{l}{\textbf{Gold Answer:} Weston-super-Mare} \\
    \hline
    \multicolumn{2}{l}{\textbf{Naive RAG}} \\
    \hline
    Retrieved & [1] Aylwin is a 1920 British silent drama film directed by Henry Edwards... \\
              & [2] Jim Wynorski (born August 14, 1950) is an American screenwriter... \\
              & [3] Frank Launder (28 January 1906 -- 23 February 1997) was a British writer, film director... \\
    \hline
    Generated & \textcolor{red}{\ding{55}} Henry Edwards \\
    \hline
    \multicolumn{2}{l}{\textbf{IndexRAG}} \\
    \hline
    Retrieved & [1] Aylwin is a 1920 British silent drama film directed by Henry Edwards... \textit{(AKU)} \\
              & [2] Henry Edwards directed both Aylwin and In the Soup, showcasing his career... \textit{(bridging fact)} \\
              & [3] \colorbox{green!15}{Henry Edwards, born on 18 September 1882 in Weston-super-Mare...} \textit{(bridging fact)} \\
    \hline
    Generated & \textcolor{green!50!black}{\ding{51}} Weston-super-Mare \\
    \hline
  \end{tabular}
  \caption{Case study from 2WikiMultiHopQA, analyzed in Section~\ref{sec:case-study}. Only the top-3 results are shown, with the decisive fact highlighted in \colorbox{green!15}{green}.}
  \label{tab:case-study}
\end{table*}

\paragraph{Multi-call methods.}
When combined with IRCoT, IndexRAG achieves the best average performance among all evaluated methods. IRCoT + IndexRAG reaches an average F1 of 55.0, slightly exceeding HippoRAG2 (54.1) and well above standalone IRCoT (41.6). Adding bridging facts to IRCoT improves performance by 13.4 points over IRCoT alone on average, with the largest gain on 2WikiMultiHopQA (+18.1), suggesting that index-time bridging facts complement query-time iterative retrieval in this setting. HippoRAG2 achieves the highest single-dataset score on HotpotQA (70.5), but requires LLM-based triple filtering and Personalized PageRank traversal at query time, resulting in significantly higher retrieval latency, as detailed in Section~\ref{sec:efficiency}.

\subsubsection{Quality-Efficiency Comparison}\label{sec:efficiency}
% ============================================================
% TABLE 2: Efficiency (MuSiQue)

Table~\ref{tab:efficiency} presents the quality-efficiency trade-off on MuSiQue. Among single-call methods, IndexRAG achieves the best EM (22.4) with a retrieval latency of 0.30 seconds, nearly identical to Naive RAG while improving EM by 3.4 points. FastGraphRAG, despite requiring only a single LLM call, requires 2.55 seconds per query, 8.5$\times$ slower than IndexRAG while achieving lower EM (17.5).
Among multi-call methods, HippoRAG2 achieves a slightly higher EM of 23.8 but at the cost of two LLM calls and a retrieval latency of 3.13 seconds, over 10$\times$ that of IndexRAG. IRCoT + IndexRAG achieves the best EM of 24.6 with a retrieval latency of 1.08 seconds, still 2.9$\times$ faster than HippoRAG2 at the retrieval stage, although it averages 3.2 LLM calls per query and we do not measure end-to-end latency including generation.

These results show that IndexRAG achieves strong performance with minimal online overhead by shifting part of the computation to indexing. To make this trade-off explicit, Appendix~\ref{app:end-to-end-cost} reports indexing cost and storage overhead. Compared with HippoRAG2, IndexRAG produces a 17--20$\times$ smaller index across the three datasets and reduces offline indexing cost by 30--50\%. Within IndexRAG, Stage~2 (bridging fact generation) contributes only 14--18\% of the total offline cost.

\subsection{Qualitative Results}

\subsubsection{Case Study}
\label{sec:case-study}

To illustrate how bridging facts improve multi-hop retrieval, we examine a concrete example from 2WikiMultiHopQA (Table~\ref{tab:case-study}).
The query ``Where was the director of the film Aylwin born?'' requires two-hop reasoning: (1) identifying the director of the film Aylwin as Henry Edwards, and (2) determining his birthplace. The relevant information is spread across two documents, requiring cross-document reasoning to connect the film to the director's biographical details.

For both methods, we present the top-3 retrieved results. Naive RAG retrieves the correct film passage as its top-ranked result, which identifies Henry Edwards as the director of Aylwin. However, the remaining two passages contain unrelated directors. The passage containing Edwards' birthplace exists in the vector store but falls outside the top-$k$, causing the LLM to output the director's name rather than his birthplace.
\begin{table}[t]
  \centering
  \begin{tabular}{lccc}
    \hline
    \textbf{Method} & \textbf{EM} & \textbf{Acc} & \textbf{F1} \\
    \hline
    Chunking & 50.2 & 54.0 & 63.6 \\
    Summary & 51.7 & 55.4 & 66.0 \\
    \textbf{QA extraction} & \textbf{53.2} & \textbf{57.4} & \textbf{67.2} \\
    \hline
  \end{tabular}
  \caption{Stage 1 extraction method comparison (\%) on HotpotQA without bridging facts.}
  \label{tab:stage1-comparison}
\end{table}
IndexRAG retrieves the same film passage as its top-ranked result. However, the remaining two slots are filled by bridging facts generated during offline indexing: one connecting Edwards to his directing career and another linking him to his birthplace in Weston-super-Mare. With this cross-document context directly available, the LLM correctly answers the query.

This example demonstrates that bridging facts do not merely summarize existing content but create new retrievable units that connect information across document boundaries, making evidence that fell outside the top-$k$ directly accessible through standard vector search.

\subsection{Ablation Study}

\subsubsection{Effect of Each Stage}\label{sec:stage_1}
% ============================================================
% TABLE 4: Stage 2 Independence (MuSiQue)
% ============================================================

\begin{table}[t]
  \centering
  \begin{tabular}{lccc}
    \hline
    \textbf{Configuration} & \textbf{EM} & \textbf{Acc} & \textbf{F1} \\
    \hline
    Naive RAG & 19.0 & 20.4 & 29.9 \\
    + Stage 2 & \textbf{22.3} & \textbf{24.4} & \textbf{34.4} \\
    \hline
    QA Extraction & 18.1 & 20.1 & 30.1 \\
    + Stage 2 & \textbf{22.4} & \textbf{24.4} &\textbf{34.4} \\
    \hline
  \end{tabular}
  \caption{Stage 2 bridging facts added on top of two different Stage 1 representations (\%) on MuSiQue.}
  \label{tab:stage2-independence}
\end{table}

\paragraph{Stage 1: Extraction Method.}
Table~\ref{tab:stage1-comparison} compares three Stage 1
strategies on HotpotQA without bridging facts. Naive chunking
yields the lowest performance with an F1 of 63.6. 
Summarization provides a modest improvement (+2.4) over 
naive chunking by compressing each document before indexing. 
QA extraction achieves the best results across all metrics,
improving F1 by 3.6 over chunking and 1.2 over summarization.
Structuring document content as question-answer pairs appears
to yield denser, more query-aligned units. Each AKU encodes
one piece of answerable information rather than raw document
text. QA extraction is therefore our default Stage 1 strategy,
used throughout unless otherwise specified.

\paragraph{Stage 2: Compatibility with the Base Representation.}
Table~\ref{tab:stage2-independence} tests whether the gain from Stage~2
depends on the Stage 1 extraction method. On MuSiQue,
bridging facts added to Naive RAG and to QA extraction reach
nearly the same final score (22.3 and 22.4 EM, respectively), so the improvement
appears independent of which Stage 1 representation is in the index. This is
practically important, since Stage 2 can be applied on top of an existing RAG
system without modifying the underlying indexing pipeline.

\begin{table}[t]
  \centering
  \small
  \begin{tabular}{llcc}
    \hline
    \textbf{Dataset} & \textbf{Datastore} & \textbf{Recall@10} & \textbf{EM} \\
    \hline
    \multirow{2}{*}{HotpotQA} & QA extraction & 66.4 & 53.2 \\
    & + Bridging & 57.5 & \textbf{53.9} \\
    \hline
    \multirow{2}{*}{2Wiki} & QA extraction & 50.5 & 41.7 \\
    & + Bridging & 46.7 & \textbf{44.8} \\
    \hline
     \multirow{2}{*}{MuSiQue} & QA extraction & 62.0 & 18.1  \\
    & + Bridging & 52.8 & \textbf{22.4} \\
    \hline
  \end{tabular}
  \caption{Recall@10 vs EM (\%) with and without bridging facts.}
  \label{tab:recall-vs-em}
\end{table}

\subsubsection{Recall vs.\ End-to-End Performance}
Table~\ref{tab:recall-vs-em} examines the relationship between 
passage recall and QA performance when bridging facts are added 
to the vector store. Across all three datasets, adding bridging 
facts consistently reduces Recall@10 yet improves EM, with the 
largest EM gain on MuSiQue (+4.3), followed by 2WikiMultiHopQA 
(+3.1) and HotpotQA (+0.7), suggesting bridging facts are most beneficial for harder datasets requiring more cross-document reasoning.

The recall drops because bridging facts compete with original 
passages for retrieval slots. Since bridging facts are newly 
generated by the LLM, they do not correspond to any original 
passage and therefore contribute no gain to passage-level recall. 
Despite this, the consistent EM improvement across all three 
datasets suggests that the cross-document reasoning encoded in 
bridging facts is more valuable for answering multi-hop questions 
than the original passages they replace.
Appendix~\ref{app:mechanism} quantifies this. When bridging facts are present in the index but withheld from the prompt, F1 stays within 1 point of Naive RAG. Adding them to the prompt raises F1 by 4.5 to 5.3 points.

\subsubsection{Performance Across Question Types}\label{sec:question_type}
\begin{figure}[t]
\centering
\includegraphics[width=0.88\columnwidth, trim=5 8 10 1, clip]{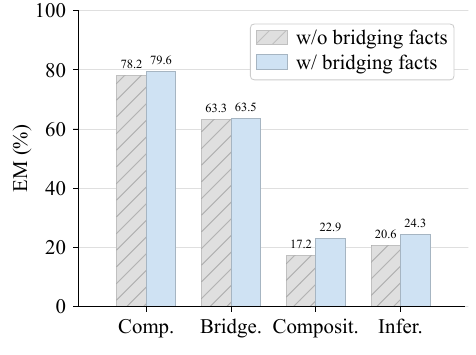}
\vspace{4pt}
\caption{Fine-grained EM (\%) on 2WikiMultiHopQA by question type. Results are shown with and without bridging facts.}
\label{fig:fine-grained}
\end{figure}
% ============================================================
% TABLE 5: 2Wiki Fine-Grained Analysis by Question Type
% ============================================================

We break results down by question type to see which kinds of reasoning benefit. Of our three datasets, 2WikiMultiHopQA has the most fine-grained type labels, so Figure~\ref{fig:fine-grained} reports its four types (comparison, bridge comparison, compositional, and inference). The largest gains from bridging facts appear on compositional (+5.7) and inference (+3.7) questions, both of which require combining evidence from separate documents to derive an answer. This aligns with the design of bridging facts, which explicitly encode cross-document connections.

Comparison questions also benefit from bridging facts (+1.4), as they often require retrieving information of two entities from different documents. Bridge comparison questions show minimal improvement (+0.2). These questions require two reasoning chains in parallel, each resolving a piece of information (e.g., birthdate) for a different entity before the two are compared. For instance, ``Which film has the director born later, X or Y?'' requires resolving the director and birthdate for both X and Y separately. Since our bridging facts are generated along sequential reasoning paths, they tend to cover only one of the two parallel paths, providing limited benefit for this question type.

These results suggest that bridging facts are most effective for question types that require synthesizing evidence across documents, such as compositional and inference questions. The relatively smaller gains on comparison and bridge comparison questions, which together account for 44\% of the dataset, partially explain why IndexRAG does not lead the single-call methods on 2WikiMultiHopQA.

\section{Conclusion}
We present \textbf{IndexRAG}, a novel approach that shifts cross-document 
reasoning from online inference to offline indexing. During indexing, IndexRAG generates bridging facts that capture 
entity-level connections across documents, stored alongside AKUs in a unified vector store. At inference time, a single 
retrieval pass and a single LLM call suffice to produce the final answer.
Extensive experiments on three multi-hop QA benchmarks show that IndexRAG achieves
the best average F1 among single-call methods, outperforming the
strongest baseline by 2.3 F1 points. The cross-document work happens once at
index time, so the online cost stays that of flat vector search. Retrieval is
6.6--10.4$\times$ faster than HippoRAG2, on an index 17--20$\times$ smaller and
cheaper to build.
IndexRAG also composes with iterative retrieval. Adding bridging facts to IRCoT raises its
average F1 to 55.0, the best result among all evaluated methods. An ablation
shows that bridging facts help mainly as direct evidence read by the generator
rather than as hints that reorder retrieval.

Future work could explore bridging fact generation conditioned on question
type to handle diverse reasoning patterns, and higher-order bridging structures
that compose several connections for longer reasoning chains.
Adapting $k_b$ per query and predicting which bridge entities are worth
expanding could further improve quality and reduce indexing cost. More broadly,
other reasoning that does not depend on the query could move to index time.

\newpage
\section*{Limitations}
IndexRAG targets multi-hop questions whose evidence can be connected through
shared entities, and its benefits are concentrated there. Gains are smaller on
comparison-style questions~\cite{ho2020constructing}, which call for parallel
lookup rather than a hidden intermediate entity
(Section~\ref{sec:question_type}), and they shrink with hop
depth, becoming negligible at four hops (Appendix~\ref{app:hop-depth}). Outside
the multi-hop regime the benefit is small: on single-hop Natural Questions \cite{kwiatkowski-etal-2019-natural}
bridging facts add 0.1 F1 over an index of AKUs alone
(Appendix~\ref{app:single-hop}).

IndexRAG also relies on LLM-based preprocessing during offline indexing,
including bridge entity extraction and bridging fact generation. Missed or noisy
entities and hallucinated bridging facts may introduce irrelevant retrieval units
and hurt performance. Dedicated NER models, source attribution, or verification
mechanisms could improve robustness. Our evaluation is limited to English
multi-hop QA benchmarks under a pre-indexed corpus setting with clean text
queries. On these same benchmarks, multi-hop retrieval has been found to amplify
rather than absorb upstream transcription errors~\cite{bao2026better}, so
robustness to noisy input is an open direction. Generalization to other
languages, domains, and dynamic corpora also remains unexplored.

\bibliography{main}

@inproceedings{ho2020constructing,
    title = "Constructing A Multi-hop {QA} Dataset for Comprehensive Evaluation of Reasoning Steps",
    author = "Ho, Xanh  and
      Duong Nguyen, Anh-Khoa  and
      Sugawara, Saku  and
      Aizawa, Akiko",
    editor = "Scott, Donia  and
      Bel, Nuria  and
      Zong, Chengqing",
    booktitle = "Proceedings of the 28th International Conference on Computational Linguistics",
    month = dec,
    year = "2020",
    address = "Barcelona, Spain (Online)",
    publisher = "International Committee on Computational Linguistics",
    url = "https://aclanthology.org/2020.coling-main.580/",
    doi = "10.18653/v1/2020.coling-main.580",
    pages = "6609--6625"
}

@inproceedings{gutierrez2024hipporag,
 author = {Guti\'{e}rrez, Bernal Jim\'{e}nez and Shu, Yiheng and Gu, Yu and Yasunaga, Michihiro and Su, Yu},
 booktitle = {Advances in Neural Information Processing Systems},
 doi = {10.52202/079017-1902},
 editor = {A. Globerson and L. Mackey and D. Belgrave and A. Fan and U. Paquet and J. Tomczak and C. Zhang},
 pages = {59532--59569},
 publisher = {Curran Associates, Inc.},
 title = {HippoRAG: Neurobiologically Inspired Long-Term Memory for Large Language Models},
 url = {https://proceedings.neurips.cc/paper_files/paper/2024/file/6ddc001d07ca4f319af96a3024f6dbd1-Paper-Conference.pdf},
 volume = {37},
 year = {2024}
}

@inproceedings{trivedi2023interleaving,
    title = "Interleaving Retrieval with Chain-of-Thought Reasoning for Knowledge-Intensive Multi-Step Questions",
    author = "Trivedi, Harsh  and
      Balasubramanian, Niranjan  and
      Khot, Tushar  and
      Sabharwal, Ashish",
    editor = "Rogers, Anna  and
      Boyd-Graber, Jordan  and
      Okazaki, Naoaki",
    booktitle = "Proceedings of the 61st Annual Meeting of the Association for Computational Linguistics (Volume 1: Long Papers)",
    month = jul,
    year = "2023",
    address = "Toronto, Canada",
    publisher = "Association for Computational Linguistics",
    url = "https://aclanthology.org/2023.acl-long.557/",
    doi = "10.18653/v1/2023.acl-long.557",
    pages = "10014--10037"
}

@article{douze2025faiss,
  author={Douze, Matthijs and Guzhva, Alexandr and Deng, Chengqi and Johnson, Jeff and Szilvasy, Gergely and Mazaré, Pierre-Emmanuel and Lomeli, Maria and Hosseini, Lucas and Jégou, Hervé},
  journal={IEEE Transactions on Big Data}, 
  title={The Faiss Library}, 
  year={2026},
  volume={12},
  number={2},
  pages={346-361},
  doi={10.1109/TBDATA.2025.3618474}}

@inproceedings{robertson1994some,
author = {Robertson, S. E. and Walker, S.},
title = {Some simple effective approximations to the 2-Poisson model for probabilistic weighted retrieval},
year = {1994},
isbn = {038719889X},
publisher = {Springer-Verlag},
address = {Berlin, Heidelberg},
booktitle = {Proceedings of the 17th Annual International ACM SIGIR Conference on Research and Development in Information Retrieval},
pages = {232–241},
numpages = {10},
location = {Dublin, Ireland},
series = {SIGIR '94}
}

@misc{edge2024local,
      title={From Local to Global: A Graph RAG Approach to Query-Focused Summarization}, 
      author={Darren Edge and Ha Trinh and Newman Cheng and Joshua Bradley and Alex Chao and Apurva Mody and Steven Truitt and Dasha Metropolitansky and Robert Osazuwa Ness and Jonathan Larson},
      year={2025},
      eprint={2404.16130},
      archivePrefix={arXiv},
      primaryClass={cs.CL},
      url={https://arxiv.org/abs/2404.16130}, 
}

@inproceedings{vaswani2017attention,
 author = {Vaswani, Ashish and Shazeer, Noam and Parmar, Niki and Uszkoreit, Jakob and Jones, Llion and Gomez, Aidan N and Kaiser, \L ukasz and Polosukhin, Illia},
 booktitle = {Advances in Neural Information Processing Systems},
 editor = {I. Guyon and U. Von Luxburg and S. Bengio and H. Wallach and R. Fergus and S. Vishwanathan and R. Garnett},
 pages = {},
 publisher = {Curran Associates, Inc.},
 title = {Attention is All you Need},
 url = {https://proceedings.neurips.cc/paper_files/paper/2017/file/3f5ee243547dee91fbd053c1c4a845aa-Paper.pdf},
 volume = {30},
 year = {2017}
}

@inproceedings{dosovitskiy2021image,
title={An Image is Worth 16x16 Words: Transformers for Image Recognition at Scale},
author={Alexey Dosovitskiy and Lucas Beyer and Alexander Kolesnikov and Dirk Weissenborn and Xiaohua Zhai and Thomas Unterthiner and Mostafa Dehghani and Matthias Minderer and Georg Heigold and Sylvain Gelly and Jakob Uszkoreit and Neil Houlsby},
booktitle={International Conference on Learning Representations},
year={2021},
url={https://openreview.net/forum?id=YicbFdNTTy}
}

@article{huang2025survey,
author = {Huang, Lei and Yu, Weijiang and Ma, Weitao and Zhong, Weihong and Feng, Zhangyin and Wang, Haotian and Chen, Qianglong and Peng, Weihua and Feng, Xiaocheng and Qin, Bing and Liu, Ting},
title = {A Survey on Hallucination in Large Language Models: Principles, Taxonomy, Challenges, and Open Questions},
year = {2025},
issue_date = {March 2025},
publisher = {Association for Computing Machinery},
address = {New York, NY, USA},
volume = {43},
number = {2},
issn = {1046-8188},
url = {https://doi.org/10.1145/3703155},
doi = {10.1145/3703155},
journal = {ACM Trans. Inf. Syst.},
month = jan,
articleno = {42},
numpages = {55}
}

@inproceedings{lewis2020retrieval,
 author = {Lewis, Patrick and Perez, Ethan and Piktus, Aleksandra and Petroni, Fabio and Karpukhin, Vladimir and Goyal, Naman and K\"{u}ttler, Heinrich and Lewis, Mike and Yih, Wen-tau and Rockt\"{a}schel, Tim and Riedel, Sebastian and Kiela, Douwe},
 booktitle = {Advances in Neural Information Processing Systems},
 editor = {H. Larochelle and M. Ranzato and R. Hadsell and M.F. Balcan and H. Lin},
 pages = {9459--9474},
 publisher = {Curran Associates, Inc.},
 title = {Retrieval-Augmented Generation for Knowledge-Intensive NLP Tasks},
 url = {https://proceedings.neurips.cc/paper_files/paper/2020/file/6b493230205f780e1bc26945df7481e5-Paper.pdf},
 volume = {33},
 year = {2020}
}

@inproceedings{sarthi2024raptor,
title={{RAPTOR}: Recursive Abstractive Processing for Tree-Organized Retrieval},
author={Parth Sarthi and Salman Abdullah and Aditi Tuli and Shubh Khanna and Anna Goldie and Christopher D Manning},
booktitle={The Twelfth International Conference on Learning Representations},
year={2024},
url={https://openreview.net/forum?id=GN921JHCRw}
}

@inproceedings{karpukhin2020dense,
    title = "Dense Passage Retrieval for Open-Domain Question Answering",
    author = "Karpukhin, Vladimir  and
      Oguz, Barlas  and
      Min, Sewon  and
      Lewis, Patrick  and
      Wu, Ledell  and
      Edunov, Sergey  and
      Chen, Danqi  and
      Yih, Wen-tau",
    editor = "Webber, Bonnie  and
      Cohn, Trevor  and
      He, Yulan  and
      Liu, Yang",
    booktitle = "Proceedings of the 2020 Conference on Empirical Methods in Natural Language Processing (EMNLP)",
    month = nov,
    year = "2020",
    address = "Online",
    publisher = "Association for Computational Linguistics",
    url = "https://aclanthology.org/2020.emnlp-main.550/",
    doi = "10.18653/v1/2020.emnlp-main.550",
    pages = "6769--6781"
}

@inproceedings{gao2023precise,
    title = "Precise Zero-Shot Dense Retrieval without Relevance Labels",
    author = "Gao, Luyu  and
      Ma, Xueguang  and
      Lin, Jimmy  and
      Callan, Jamie",
    editor = "Rogers, Anna  and
      Boyd-Graber, Jordan  and
      Okazaki, Naoaki",
    booktitle = "Proceedings of the 61st Annual Meeting of the Association for Computational Linguistics (Volume 1: Long Papers)",
    month = jul,
    year = "2023",
    address = "Toronto, Canada",
    publisher = "Association for Computational Linguistics",
    url = "https://aclanthology.org/2023.acl-long.99/",
    doi = "10.18653/v1/2023.acl-long.99",
    pages = "1762--1777"
}

@inproceedings{press2023measuring,
    title = "Measuring and Narrowing the Compositionality Gap in Language Models",
    author = "Press, Ofir  and
      Zhang, Muru  and
      Min, Sewon  and
      Schmidt, Ludwig  and
      Smith, Noah  and
      Lewis, Mike",
    editor = "Bouamor, Houda  and
      Pino, Juan  and
      Bali, Kalika",
    booktitle = "Findings of the Association for Computational Linguistics: EMNLP 2023",
    month = dec,
    year = "2023",
    address = "Singapore",
    publisher = "Association for Computational Linguistics",
    url = "https://aclanthology.org/2023.findings-emnlp.378/",
    doi = "10.18653/v1/2023.findings-emnlp.378",
    pages = "5687--5711"
}

@inproceedings{brown2020language,
 author = {Brown, Tom and Mann, Benjamin and Ryder, Nick and Subbiah, Melanie and Kaplan, Jared D and Dhariwal, Prafulla and Neelakantan, Arvind and Shyam, Pranav and Sastry, Girish and Askell, Amanda and Agarwal, Sandhini and Herbert-Voss, Ariel and Krueger, Gretchen and Henighan, Tom and Child, Rewon and Ramesh, Aditya and Ziegler, Daniel and Wu, Jeffrey and Winter, Clemens and Hesse, Chris and Chen, Mark and Sigler, Eric and Litwin, Mateusz and Gray, Scott and Chess, Benjamin and Clark, Jack and Berner, Christopher and McCandlish, Sam and Radford, Alec and Sutskever, Ilya and Amodei, Dario},
 booktitle = {Advances in Neural Information Processing Systems},
 editor = {H. Larochelle and M. Ranzato and R. Hadsell and M.F. Balcan and H. Lin},
 pages = {1877--1901},
 publisher = {Curran Associates, Inc.},
 title = {Language Models are Few-Shot Learners},
 url = {https://proceedings.neurips.cc/paper_files/paper/2020/file/1457c0d6bfcb4967418bfb8ac142f64a-Paper.pdf},
 volume = {33},
 year = {2020}
}

@InProceedings{pmlr-v267-gutierrez25a,
  title = {From {RAG} to Memory: Non-Parametric Continual Learning for Large Language Models},
  author = {Guti{\'e}rrez, Bernal Jim{\'e}nez and Shu, Yiheng and Qi, Weijian and Zhou, Sizhe and Su, Yu},
  booktitle = {Proceedings of the 42nd International Conference on Machine Learning},
  pages = {21497--21515},
  year = {2025},
  editor = {Singh, Aarti and Fazel, Maryam and Hsu, Daniel and Lacoste-Julien, Simon and Berkenkamp, Felix and Maharaj, Tegan and Wagstaff, Kiri and Zhu, Jerry},
  volume = {267},
  series = {Proceedings of Machine Learning Research},
  month = {13--19 Jul},
  publisher = {PMLR},
  url = {https://proceedings.mlr.press/v267/gutierrez25a.html}
}

@inbook{chhikara2025mem0,
  title={{Mem0}: Building Production-Ready {AI} Agents with Scalable Long-Term Memory},
  ISBN={9781643686318},
  ISSN={1879-8314},
  url={http://dx.doi.org/10.3233/FAIA251160},
  DOI={10.3233/faia251160},
  booktitle={ECAI 2025},
  series={Frontiers in Artificial Intelligence and Applications},
  volume={413},
  pages={2993--3000},
  publisher={IOS Press},
  author={Chhikara, Prateek and Khant, Dev and Aryan, Saket and Singh, Taranjeet and Yadav, Deshraj},
  year={2025},
  month=Oct
}

@inproceedings{kang2025memory,
    title = "Memory {OS} of {AI} Agent",
    author = "Kang, Jiazheng  and
      Ji, Mingming  and
      Zhao, Zhe  and
      Bai, Ting",
    editor = "Christodoulopoulos, Christos  and
      Chakraborty, Tanmoy  and
      Rose, Carolyn  and
      Peng, Violet",
    booktitle = "Proceedings of the 2025 Conference on Empirical Methods in Natural Language Processing",
    month = nov,
    year = "2025",
    address = "Suzhou, China",
    publisher = "Association for Computational Linguistics",
    url = "https://aclanthology.org/2025.emnlp-main.1318/",
    doi = "10.18653/v1/2025.emnlp-main.1318",
    pages = "25961--25970",
    ISBN = "979-8-89176-332-6"
}

@inproceedings{thakur2021beir,
 author = {Thakur, Nandan and Reimers, Nils and R\"{u}ckl\'{e}, Andreas and Srivastava, Abhishek and Gurevych, Iryna},
 booktitle = {Proceedings of the Neural Information Processing Systems Track on Datasets and Benchmarks},
 editor = {J. Vanschoren and S. Yeung},
 title = {{BEIR}: A Heterogeneous Benchmark for Zero-shot Evaluation of Information Retrieval Models},
 url = {https://datasets-benchmarks-proceedings.neurips.cc/paper_files/paper/2021/file/65b9eea6e1cc6bb9f0cd2a47751a186f-Paper-round2.pdf},
 volume = {1},
 year = {2021}
}

@misc{bao2026better,
      title={Better Retrieval, Worse Robustness: How Multi-hop RAG Amplifies Upstream ASR Errors}, 
      author={Zhenghua Bao},
      year={2026},
      eprint={2608.22872},
      archivePrefix={arXiv},
      primaryClass={cs.CL},
      url={https://arxiv.org/abs/2608.22872}, 
}

@misc{defossez2024moshi,
      title={Moshi: a speech-text foundation model for real-time dialogue}, 
      author={Alexandre Défossez and Laurent Mazaré and Manu Orsini and Amélie Royer and Patrick Pérez and Hervé Jégou and Edouard Grave and Neil Zeghidour},
      year={2024},
      eprint={2410.00037},
      archivePrefix={arXiv},
      primaryClass={eess.AS},
      url={https://arxiv.org/abs/2410.00037}, 
}

@inproceedings{zhang-etal-2023-speechgpt,
    title = "{S}peech{GPT}: Empowering Large Language Models with Intrinsic Cross-Modal Conversational Abilities",
    author = "Zhang, Dong  and
      Li, Shimin  and
      Zhang, Xin  and
      Zhan, Jun  and
      Wang, Pengyu  and
      Zhou, Yaqian  and
      Qiu, Xipeng",
    editor = "Bouamor, Houda  and
      Pino, Juan  and
      Bali, Kalika",
    booktitle = "Findings of the Association for Computational Linguistics: EMNLP 2023",
    month = dec,
    year = "2023",
    address = "Singapore",
    publisher = "Association for Computational Linguistics",
    url = "https://aclanthology.org/2023.findings-emnlp.1055/",
    doi = "10.18653/v1/2023.findings-emnlp.1055",
    pages = "15757--15773"
}

@article{JMLR:v21:20-074,
  author  = {Colin Raffel and Noam Shazeer and Adam Roberts and Katherine Lee and Sharan Narang and Michael Matena and Yanqi Zhou and Wei Li and Peter J. Liu},
  title   = {Exploring the Limits of Transfer Learning with a Unified Text-to-Text Transformer},
  journal = {Journal of Machine Learning Research},
  year    = {2020},
  volume  = {21},
  number  = {140},
  pages   = {1--67},
  url     = {http://jmlr.org/papers/v21/20-074.html}
}

@inproceedings{you-etal-2025-ms,
    title = "{MS}-{RAG}: Simple and Effective Multi-Semantic Retrieval-Augmented Generation",
    author = "You, Xiaozhou  and
      Luo, Yahui  and
      Gu, Lihong",
    editor = "Christodoulopoulos, Christos  and
      Chakraborty, Tanmoy  and
      Rose, Carolyn  and
      Peng, Violet",
    booktitle = "Proceedings of the 2025 Conference on Empirical Methods in Natural Language Processing",
    month = nov,
    year = "2025",
    address = "Suzhou, China",
    publisher = "Association for Computational Linguistics",
    url = "https://aclanthology.org/2025.emnlp-main.1151/",
    doi = "10.18653/v1/2025.emnlp-main.1151",
    pages = "22609--22625",
    ISBN = "979-8-89176-332-6"
}

@inproceedings{NEURIPS2025_19909c36,
 author = {Xu, Wujiang and Liang, Zujie and Mei, Kai and Gao, Hang and Tan, Juntao and Zhang, Yongfeng},
 booktitle = {Advances in Neural Information Processing Systems},
 doi = {10.52202/085713-0593},
 editor = {D. Belgrave and C. Zhang and H. Lin and R. Pascanu and P. Koniusz and M. Ghassemi and N. Chen},
 pages = {17577--17604},
 publisher = {Curran Associates, Inc.},
 title = {A-Mem: Agentic Memory for LLM Agents},
 url = {https://proceedings.neurips.cc/paper_files/paper/2025/file/19909c36f51abc4856b4560aff3d36d6-Paper-Conference.pdf},
 volume = {38, Main Conference},
 year = {2025}
}

@inproceedings{yang-etal-2018-hotpotqa,
    title = "{H}otpot{QA}: A Dataset for Diverse, Explainable Multi-hop Question Answering",
    author = "Yang, Zhilin  and
      Qi, Peng  and
      Zhang, Saizheng  and
      Bengio, Yoshua  and
      Cohen, William  and
      Salakhutdinov, Ruslan  and
      Manning, Christopher D.",
    editor = "Riloff, Ellen  and
      Chiang, David  and
      Hockenmaier, Julia  and
      Tsujii, Jun{'}ichi",
    booktitle = "Proceedings of the 2018 Conference on Empirical Methods in Natural Language Processing",
    month = oct # "-" # nov,
    year = "2018",
    address = "Brussels, Belgium",
    publisher = "Association for Computational Linguistics",
    url = "https://aclanthology.org/D18-1259/",
    doi = "10.18653/v1/D18-1259",
    pages = "2369--2380"
}

@article{trivedi-etal-2022-musique,
    title = "{M}u{S}i{Q}ue: Multihop Questions via Single-hop Question Composition",
    author = "Trivedi, Harsh  and
      Balasubramanian, Niranjan  and
      Khot, Tushar  and
      Sabharwal, Ashish",
    editor = "Roark, Brian  and
      Nenkova, Ani",
    journal = "Transactions of the Association for Computational Linguistics",
    volume = "10",
    year = "2022",
    address = "Cambridge, MA",
    publisher = "MIT Press",
    url = "https://aclanthology.org/2022.tacl-1.31/",
    doi = "10.1162/tacl_a_00475",
    pages = "539--554"
}

@inproceedings{rajpurkar-etal-2016-squad,
    title = "{SQ}u{AD}: 100,000+ Questions for Machine Comprehension of Text",
    author = "Rajpurkar, Pranav  and
      Zhang, Jian  and
      Lopyrev, Konstantin  and
      Liang, Percy",
    editor = "Su, Jian  and
      Duh, Kevin  and
      Carreras, Xavier",
    booktitle = "Proceedings of the 2016 Conference on Empirical Methods in Natural Language Processing",
    month = nov,
    year = "2016",
    address = "Austin, Texas",
    publisher = "Association for Computational Linguistics",
    url = "https://aclanthology.org/D16-1264/",
    doi = "10.18653/v1/D16-1264",
    pages = "2383--2392"
}

@inproceedings{chen-etal-2024-dense,
    title = "Dense {X} Retrieval: What Retrieval Granularity Should We Use?",
    author = "Chen, Tong  and
      Wang, Hongwei  and
      Chen, Sihao  and
      Yu, Wenhao  and
      Ma, Kaixin  and
      Zhao, Xinran  and
      Zhang, Hongming  and
      Yu, Dong",
    editor = "Al-Onaizan, Yaser  and
      Bansal, Mohit  and
      Chen, Yun-Nung",
    booktitle = "Proceedings of the 2024 Conference on Empirical Methods in Natural Language Processing",
    month = nov,
    year = "2024",
    address = "Miami, Florida, USA",
    publisher = "Association for Computational Linguistics",
    url = "https://aclanthology.org/2024.emnlp-main.845/",
    doi = "10.18653/v1/2024.emnlp-main.845",
    pages = "15159--15177"
}

@article{kwiatkowski-etal-2019-natural,
    title = "Natural Questions: A Benchmark for Question Answering Research",
    author = "Kwiatkowski, Tom  and
      Palomaki, Jennimaria  and
      Redfield, Olivia  and
      Collins, Michael  and
      Parikh, Ankur  and
      Alberti, Chris  and
      Epstein, Danielle  and
      Polosukhin, Illia  and
      Devlin, Jacob  and
      Lee, Kenton  and
      Toutanova, Kristina  and
      Jones, Llion  and
      Kelcey, Matthew  and
      Chang, Ming-Wei  and
      Dai, Andrew M.  and
      Uszkoreit, Jakob  and
      Le, Quoc  and
      Petrov, Slav",
    editor = "Lee, Lillian  and
      Johnson, Mark  and
      Roark, Brian  and
      Nenkova, Ani",
    journal = "Transactions of the Association for Computational Linguistics",
    volume = "7",
    year = "2019",
    address = "Cambridge, MA",
    publisher = "MIT Press",
    url = "https://aclanthology.org/Q19-1026/",
    doi = "10.1162/tacl_a_00276",
    pages = "452--466"
}

@inproceedings{hu-etal-2026-evermemos,
    title = "{E}ver{M}em{OS}: A Self-Organizing Memory Operating System for Structured Long-Horizon Reasoning",
    author = "Hu, Chuanrui  and
      Gao, Xingze  and
      Zhou, Zuyi  and
      Xu, Dannong  and
      Bai, Yi  and
      Li, Xintong  and
      Zhang, Hui  and
      Li, Tong  and
      Zhang, Chong  and
      Bing, Lidong  and
      Deng, Yafeng",
    editor = "Liakata, Maria  and
      Moreira, Viviane P.  and
      Zhang, Jiajun  and
      Jurgens, David",
    booktitle = "Proceedings of the 64th Annual Meeting of the {A}ssociation for {C}omputational {L}inguistics (Volume 1: Long Papers)",
    month = jul,
    year = "2026",
    address = "San Diego, California, United States",
    publisher = "Association for Computational Linguistics",
    url = "https://aclanthology.org/2026.acl-long.2125/",
    doi = "10.18653/v1/2026.acl-long.2125",
    pages = "45836--45853",
    ISBN = "979-8-89176-390-6"
}

@InProceedings{pmlr-v139-radford21a,
  title =      {Learning Transferable Visual Models From Natural Language Supervision},
  author =       {Radford, Alec and Kim, Jong Wook and Hallacy, Chris and Ramesh, Aditya and Goh, Gabriel and Agarwal, Sandhini and Sastry, Girish and Askell, Amanda and Mishkin, Pamela and Clark, Jack and Krueger, Gretchen and Sutskever, Ilya},
  booktitle =      {Proceedings of the 38th International Conference on Machine Learning},
  pages =      {8748--8763},
  year =      {2021},
  editor =      {Meila, Marina and Zhang, Tong},
  volume =      {139},
  series =      {Proceedings of Machine Learning Research},
  month =      {18--24 Jul},
  publisher =    {PMLR},
  url =      {https://proceedings.mlr.press/v139/radford21a.html}
}

@inproceedings{xu-etal-2026-self,
    title = "Self-Correcting {RAG}: Enhancing Faithfulness via {MMKP} Context Selection and {NLI}-Guided {MCTS}",
    author = "Xu, Shijia  and
      Wu, Zhou  and
      Jia, Xiaolong  and
      Wang, Yu  and
      Liu, Kai  and
      Dong, April Xiaowen",
    editor = "Liakata, Maria  and
      Moreira, Viviane P.  and
      Zhang, Jiajun  and
      Jurgens, David",
    booktitle = "Findings of the {A}ssociation for {C}omputational {L}inguistics: {ACL} 2026",
    month = jul,
    year = "2026",
    address = "San Diego, California, United States",
    publisher = "Association for Computational Linguistics",
    url = "https://aclanthology.org/2026.findings-acl.1052/",
    doi = "10.18653/v1/2026.findings-acl.1052",
    pages = "20953--20976",
    ISBN = "979-8-89176-395-1"
}

\newpage
\appendix

\section{Evaluation Metrics}\label{app:eval_metrics}
Let $\hat{a}$ and $a^*$ denote the normalized prediction and gold answer, and let $t(\cdot)$ denote the token sequence of a string and $T(\cdot)$ its set of distinct tokens.
Exact Match (EM) measures whether the prediction is identical to the gold answer:
\begin{equation}
    \text{EM} = 
    \begin{cases}
        1 & \text{if } \hat{a} = a^* \\
        0 & \text{otherwise}
    \end{cases}
\end{equation}

Accuracy (Acc) measures whether the gold answer appears as a substring of the prediction:
\begin{equation}
    \text{Acc} = 
    \begin{cases}
        1 & \text{if } a^* \subseteq \hat{a} \\
        0 & \text{otherwise}
    \end{cases}
\end{equation}

F1 computes the token-level harmonic mean of precision and recall, counting
each matching token once in the numerator and all tokens in the denominators:
\begin{align}
    P = \frac{|T(\hat{a}) \cap T(a^*)|}{|t(\hat{a})|}, \quad & R = \frac{|T(\hat{a}) \cap T(a^*)|}{|t(a^*)|} \\[4pt]
    F_1 &= \frac{2PR}{P+R}
\end{align}

\section{Balanced Context Selection}
\label{app:algorithm}

Algorithm~\ref{alg:balanced} presents the full procedure for balanced 
context selection, which controls the proportion of bridging facts 
in the retrieved context.

\begin{algorithm}[h]
\caption{Balanced Context Selection}
\label{alg:balanced}
\begin{algorithmic}[1]
\State $C \leftarrow \emptyset$, $n_b \leftarrow 0$
\For{$r_j$ in $R$}
    \If{$|C| = k$}
        \State \textbf{break}
    \EndIf
    \If{$r_j \in \mathcal{A}$ \textbf{or} $n_b < k_b$}
        \State $C \leftarrow C \cup \{r_j\}$
        \If{$r_j \in \mathcal{B}$}
            \State $n_b \leftarrow n_b + 1$
        \EndIf
    \EndIf
\EndFor
\State \Return $C$
\end{algorithmic}
\end{algorithm}

\section{Effect of Balanced Context Selection}
% ============================================================
% TABLE 6: Balanced Retrieval Ablation
% ============================================================

The parameter $k_b$ controls the maximum number of bridging facts 
included in the retrieved context. Table~\ref{tab:balanced-ablation} 
shows results for $k_b \in \{0, 1, 2, 3, 5\}$, where $k_b = 0$ 
corresponds to retrieval without bridging facts. Performance peaks at $k_b = 2$ on HotpotQA and MuSiQue (with MuSiQue tying at $k_b = 2$ and $k_b = 3$), and at $k_b = 3$ on 2WikiMultiHopQA, then declines as $k_b$ increases further. This is because 
bridging facts are shorter than AKUs on average (166 vs.\ 634 
characters), so including too many bridging facts displaces original context, which contains more informative content. We observe that datasets requiring more cross-document reasoning, 
such as MuSiQue, tolerate a higher $k_b$.

\begin{table}[t]
  \centering
  \small
  \begin{tabular}{lccc}
    \hline
    $k_b$ & \textbf{HotpotQA} & \textbf{2Wiki} & \textbf{MuSiQue} \\
    \hline
    0 (no bridging) & 53.2 & 41.7 & 18.1 \\
    1 & 52.7 & 43.3 & 21.6 \\
    2 & \textbf{54.1} & 43.2 & 22.4 \\
    3 & 53.9 & \textbf{44.8} & \textbf{22.4} \\
    5 & 52.9 & 44.4 & 22.3 \\
    \hline
  \end{tabular}
  \caption{Balanced retrieval ablation (EM, \%).}
  \label{tab:balanced-ablation}
\end{table}

\section{End-to-End Cost Accounting}
\label{app:end-to-end-cost}

Index-time reasoning trades online cost for offline cost, so both sides need accounting. Table~\ref{tab:end-to-end-cost} reports retrieval latency, index size on disk, and offline indexing cost against HippoRAG2 on each dataset.

\begin{table}[t]
  \centering
  \small
  \setlength{\tabcolsep}{4pt}
  \begin{tabular}{lccc}
    \hline
    \textbf{Dataset} & \textbf{Latency (s)} & \textbf{Index (MB)} & \textbf{Cost (\$)} \\
    \hline
    HotpotQA & -- & 144 / 2657 & 4.43 / 6.37 \\
    2Wiki & 0.44 / 2.89 & 83 / 1426 & 2.60 / 5.22 \\
    MuSiQue & 0.30 / 3.13 & 114 / 2278 & 3.67 / 6.56 \\
    \hline
  \end{tabular}
  \caption{End-to-end cost. Each cell is IndexRAG / HippoRAG2. Latency is
  retrieval time per query, and was not measured on HotpotQA. Cost is recorded
  directly from OpenAI API responses over the full retrieval corpus of each
  dataset using
  gpt-4o-mini pricing.}
  \label{tab:end-to-end-cost}
\end{table}

IndexRAG is cheaper to index than HippoRAG2 despite generating new bridging facts. The bridging-fact generation stage accounts for only 18\% of the total IndexRAG offline cost on HotpotQA, 17\% on 2WikiMultiHopQA, and 14\% on MuSiQue, suggesting that index-time bridging adds little at the margin rather than hiding an expensive preprocessing step.

\section{Single-Hop Sanity Check}
\label{app:single-hop}

To verify that adding bridging facts does not harm retrieval quality outside the multi-hop regime IndexRAG targets, we run a sanity check on Natural Questions~\cite{kwiatkowski-etal-2019-natural}. We sample 1{,}000 answerable questions from the validation set, retrieve their associated Wikipedia articles from the DPR-preprocessed BEIR/NQ corpus~\cite{karpukhin2020dense, thakur2021beir}, and merge passages per article, resulting in 924 unique articles after deduplication. This mirrors our closed-corpus multi-hop QA protocol.

\begin{table}[t]
  \centering
  \small
  \begin{tabular}{lcc}
    \hline
    \textbf{Method} & \textbf{EM} & \textbf{F1} \\
    \hline
    Naive RAG & 33.2 & 47.7 \\
    AKU-only (no bridging) & 35.4 & 49.6 \\
    IndexRAG (AKU + bridging) & \textbf{35.9} & \textbf{49.7} \\
    \hline
  \end{tabular}
  \caption{Single-hop sanity check on Natural Questions.}
  \label{tab:nq-single-hop}
\end{table}

Table~\ref{tab:nq-single-hop} shows that IndexRAG does not degrade single-hop performance. The gain over Naive RAG comes primarily from AKU granularity rather than bridging facts: AKU-only retrieval improves EM by 2.2 points, while adding bridging facts contributes a further 0.5 EM. This is expected because bridging facts are designed for cross-document inference, whereas single-hop questions mainly benefit from more focused retrieval units.

\section{Mechanism Analysis: Direct Evidence or Retrieval Hint?}
\label{app:mechanism}

A bridging fact could help in two ways: the generator may read it as evidence, or its presence in the index may change which AKUs are retrieved. To separate the two, we evaluate a variant that keeps bridging facts in the index, where they still compete for top-$k$ slots, but removes them from the prompt before generation. If the benefit comes from better retrieval, this variant should still beat Naive RAG on F1. If the generator has to read the fact itself, that advantage should vanish.

\begin{table}[t]
  \centering
  \small
  \begin{tabular}{lcccc}
    \hline
    \multirow{2}{*}{\textbf{Method}} & \multicolumn{2}{c}{\textbf{2Wiki}} & \multicolumn{2}{c}{\textbf{MuSiQue}} \\
    & EM & F1 & EM & F1 \\
    \hline
    Naive RAG & 42.2 & 47.7 & 19.0 & 29.9 \\
    Retrieval-only & 41.7 & 47.2 & 17.7 & 29.1 \\
    Full IndexRAG & \textbf{44.8} & \textbf{51.7} & \textbf{22.4} & \textbf{34.4} \\
    \hline
  \end{tabular}
  \caption{Mechanism ablation. Bridging facts remain in the index but are removed from the final prompt in the middle row.}
  \label{tab:mechanism-ablation}
\end{table}

Table~\ref{tab:mechanism-ablation} shows the retrieval-only variant staying within 1 F1 of Naive RAG and falling far below full IndexRAG. Putting the facts back into the prompt recovers 4.5 F1 on 2WikiMultiHopQA and 5.3 F1 on MuSiQue. The generator therefore has to read the fact itself, and better retrieval alone accounts for almost none of the improvement.

The same variant also puts a ceiling on what slot displacement costs. Its F1 sits 0.5 below Naive RAG on 2WikiMultiHopQA and 0.8 below on MuSiQue. The variant differs from Naive RAG in two ways: bridging facts take up top-$k$ slots, and fixed chunks are replaced by AKUs. Those gaps therefore bound displacement rather than measure it. The AKU-only index isolates it, differing from the retrieval-only variant only in whether bridging facts are present. On MuSiQue, F1 falls from 30.1 (Table~\ref{tab:stage2-independence}) to 29.1, a drop of 1.0.

\section{Effect of Hop Depth}
\label{app:hop-depth}

We further analyze IndexRAG on MuSiQue by hop depth, using the dataset's $N_{\text{hop}}$ field to partition the 1{,}000 questions into 535 two-hop, 318 three-hop, and 147 four-hop subsets. Table~\ref{tab:hop-depth} reports the improvement of IndexRAG over the no-bridging variant on each subset.

\begin{table}[t]
  \centering
  \small
  \begin{tabular}{lccc}
    \hline
    \textbf{MuSiQue subset} & \textbf{N} & $\Delta$\textbf{EM} & $\Delta$\textbf{F1} \\
    \hline
    2-hop & 535 & +6.2 & +5.4 \\
    3-hop & 318 & +3.1 & +4.5 \\
    4-hop & 147 & +1.4 & $-$0.2 \\
    \hline
  \end{tabular}
  \caption{IndexRAG improvement by hop depth on MuSiQue.}
  \label{tab:hop-depth}
\end{table}

The benefit decreases with hop depth and is negligible on four-hop questions. This pattern is consistent with the design of bridging facts, which primarily encode pairwise connections between documents sharing an entity. Longer reasoning chains may require composing multiple bridging facts or generating higher-order bridging structures.

\section{Prompt Templates}
\label{sec:prompts}
We present all prompt templates used in our experiments.
Figure~\ref{fig:prompt-stage1} shows the Stage 1 AKU extraction prompt,
which instructs the LLM to extract factual information from a document
as atomic question-answer pairs.
Figure~\ref{fig:prompt-summary} shows the summary-based Stage 1 baseline. The LLM is asked to produce a comprehensive summary of each document.
For Stage 2, Figure~\ref{fig:prompt-stage2} provides documents sharing
a common entity and prompts the LLM to generate cross-document bridging
facts in JSON format.
Figure~\ref{fig:prompt-answer} is the answer generation prompt shared
across all methods, requiring the LLM to output concise, exact answers
without explanation.
Figure~\ref{fig:prompt-ircot} presents the IRCoT reasoning prompt,
which elicits step-by-step chain-of-thought reasoning and guides the LLM
to suggest follow-up search queries when additional information is needed.

% ---- Stage 1: AKU Extraction ----
\begin{figure*}[t]
\begin{tcolorbox}[
    colback=gray!5,
    colframe=gray!30,
    fontupper=\small,
    left=6pt, right=6pt, top=4pt, bottom=4pt,
    boxrule=0.4pt,
    width=\textwidth,
    enlarge left by=0pt,
    enlarge right by=0pt,
]
\textbf{[context]}\\
\{text\}\\
\textbf{[/context]}

You are an expert information extractor. Extract ALL factual
information from the content above as question-answer pairs.

\textbf{\# Requirements:}\\
- Each FAQ answers exactly ONE question (atomic)\\
- Questions are natural English (``Who directed X?'' not ``Director of X'')\\
- Answers are complete and self-contained\\
- Extract every entity, date, relationship, event, and property\\
- ALL output in English

\textbf{\# Output Schema:}
\begin{verbatim}
{"all_faqs": [
   {"question": "...", "answer": "...",
    "entities": ["Entity1", "Entity2"]}
]}
\end{verbatim}

\textbf{\# Example:}

Source: ``Ed Wood is a 1994 American biographical film directed by
Tim Burton, starring Johnny Depp.''
\begin{verbatim}
{"all_faqs": [
  {"question": "Who directed Ed Wood?",
   "answer": "Tim Burton directed Ed Wood.",
   "entities": ["Ed Wood", "Tim Burton"]}
]}
\end{verbatim}

Return ONLY valid JSON. Start with \texttt{\{}, end with
\texttt{\}}. No markdown, no comments.
\end{tcolorbox}
\caption{\textbf{Prompt template for Stage 1 AKU extraction.} The
\texttt{entities} field is the entity set $E_i$ used by Stage~2. The example
is shown with one of its four question-answer pairs for brevity.}
\label{fig:prompt-stage1}
\end{figure*}

% ---- Stage 1: Summary Baseline ----
\begin{figure*}[t]
\begin{tcolorbox}[
    colback=gray!5,
    colframe=gray!30,
    fontupper=\small,
    left=6pt, right=6pt, top=4pt, bottom=4pt,
    boxrule=0.4pt,
    width=\textwidth,
]
Given the following document, write a comprehensive summary that
captures all key facts, entities, relationships, and details.
Be thorough and do not omit important information.

\textbf{Document:}
\{text\}

\textbf{Summary:}
\end{tcolorbox}
\caption{\textbf{Prompt template for the Stage 1 summary strategy.}}
\label{fig:prompt-summary}
\end{figure*}

% ---- Stage 2: Bridging Fact Generation ----
\begin{figure*}[t]
\begin{tcolorbox}[
    colback=gray!5,
    colframe=gray!30,
    fontupper=\small,
    left=6pt, right=6pt, top=4pt, bottom=4pt,
    boxrule=0.4pt,
    width=\textwidth,
]
Given the following information about ``\{entity\}'' from multiple
source documents, generate bridging facts that connect information
across these documents.

\{doc\_sections\}

\textbf{Requirements:}\\
- Each bridging fact must combine information from 2+ documents\\
- Be factually accurate --- only connect information that is logically related\\
- Each fact should be self-contained and understandable without context\\
- Do not generate speculative connections\\
- If documents share the entity name but are about unrelated topics, return empty

Return a JSON array of strings. If no meaningful connections exist, return [].
\end{tcolorbox}
\caption{\textbf{Prompt template for Stage 2 bridging fact generation.}}
\label{fig:prompt-stage2}
\end{figure*}

% ---- Answer Generation ----
\begin{figure*}[t]
\begin{tcolorbox}[
    colback=gray!5,
    colframe=gray!30,
    fontupper=\small,
    left=6pt, right=6pt, top=4pt, bottom=4pt,
    boxrule=0.4pt,
    width=\textwidth,
]
You are a precise question answering assistant. Answer with ONLY
the exact information requested, with no explanations or extra words.
If the answer is a name, give only the name. If the answer is a number,
give only the number. If the answer is yes/no, give only yes or no.

\textbf{Context:}
\{context\}

\textbf{Question:}
\{question\}
\end{tcolorbox}
\caption{\textbf{Prompt template for answer generation (used by all methods).}}
\label{fig:prompt-answer}
\end{figure*}

% ---- IRCoT Reasoning ----
\begin{figure*}[t]
\begin{tcolorbox}[
    colback=gray!5,
    colframe=gray!30,
    fontupper=\small,
    left=6pt, right=6pt, top=4pt, bottom=4pt,
    boxrule=0.4pt,
    width=\textwidth,
]
You are a reasoning assistant that helps answer multi-hop questions
step by step.

\textbf{Question:} \{question\}

\textbf{Retrieved Information:}
\{context\}

\textbf{Reasoning so far:}
\{cot\_so\_far\}

Write ONE brief reasoning sentence that makes progress toward
answering the question. If more information is needed, suggest a
specific search query.

\textbf{Format your response as:}\\
Reasoning: \textlangle one sentence of reasoning\textrangle\\
Search: \textlangle next search query, or DONE if ready to answer\textrangle
\end{tcolorbox}
\caption{\textbf{Prompt template for IRCoT chain-of-thought reasoning,
following \citet{trivedi2023interleaving}.}}
\label{fig:prompt-ircot}
\end{figure*}

\end{document}